%% file: ms.tex
\documentclass[journal,twoside,web]{ieeecolor}
\pdfoutput=1

\usepackage{tmi}
\usepackage{cite}
\usepackage{amsmath,amssymb,amsfonts}
\usepackage{algorithmic}
\usepackage{graphicx}
\usepackage{textcomp}
\usepackage{booktabs}  
\usepackage{placeins}
\usepackage{bm}
\usepackage{multirow} 
\usepackage{pifont} 
\usepackage{makecell}
\usepackage{color}
\usepackage{colortbl}
\usepackage[HTML]{xcolor}
\usepackage{caption}
\usepackage{dirtytalk}
\MakeRobust{\say}
\usepackage{enumerate}
\usepackage[export]{adjustbox}
\newcommand{\cmark}{\ding{51}}

\usepackage{nomencl}
\makenomenclature

\usepackage{hyperref}
\hypersetup{pdfauthor={Philip M\"uller* and Felix Meissen* and Georgios Kaissis and Daniel Rueckert},pdftitle={Weakly Supervised Object Detection in Chest X-Rays with Differentiable ROI Proposal Networks and Soft ROI Pooling}}
\usepackage[capitalize]{cleveref}
\crefname{section}{Sec.}{Secs.}
\Crefname{section}{Section}{Sections}
\Crefname{table}{Table}{Tables}
\crefname{table}{Table}{Tables}

\newcommand{\nf}{\varnothing}

\newcommand{\nfor}{\lor\nf}
\newcommand{\nfand}{\land\nf}

\newcommand{\yc}[1][c]{y_{#1}}

\newcommand{\hp}{\bm{h}^{\mathcal{P}}_{m, n}}

\newcommand{\agghpc}[1][c]{\bar{\bm{h}}^{\mathcal{P}}_{#1}}

\newcommand{\hr}{\bm{h}^{\mathcal{R}}_{k}}

\newcommand{\agghrc}[1][c]{\bm{\bar{h}}^{\mathcal{R}}_{#1}}

\newcommand{\ppc}[1][c]{p^{\mathcal{P}}_{m, n, #1}}
\newcommand{\ppnf}{\ppc[\nf]}

\newcommand{\aggppc}[1][c]{\bar{p}^{\mathcal{P}}_{#1}}

\newcommand{\prc}[1][c]{p^{\mathcal{R}}_{k, #1}}
\newcommand{\prnf}{\prc[\nf]}

\newcommand{\aggprc}[1][c]{\bar{p}^{\mathcal{R}}_{#1}}
\newcommand{\aggprnfor}{\aggprc[\nfor]}
\newcommand{\aggprnfand}{\aggprc[\nfand]}

\newcommand{\loss}[1]{\mathcal{L}_{\text{#1}}}
\newcommand{\lossp}[1]{\loss{#1}^{\mathcal{P}}}
\newcommand{\lossr}[1]{\loss{#1}^{\mathcal{R}}}
\newcommand{\losspr}{\loss{}^{\mathcal{P}\leftrightarrow\mathcal{R}}}

\DeclareMathOperator*{\noisyor}{noisyOR}
\DeclareMathOperator*{\noisyand}{noisyAND}
\DeclareMathOperator*{\lse}{LSE}
\DeclareMathOperator*{\argmax}{arg\,max}
\DeclareMathOperator*{\kl}{D_\text{KL}}

\definecolor{Gray}{gray}{0.9}
\definecolor{DarkGray}{gray}{0.5}
\definecolor{greenish}{HTML}{2ca02c}
\definecolor{yellowish}{HTML}{bcbd22}
\definecolor{redish}{HTML}{d62728}

\newcommand{\std}[1]{\textcolor[gray]{0.3}{\scriptsize$\pm$#1}}

\def\BibTeX{{\rm B\kern-.05em{\sc i\kern-.025em b}\kern-.08em
    T\kern-.1667em\lower.7ex\hbox{E}\kern-.125emX}}

\begin{document}
    \title{Weakly Supervised Object Detection in Chest X-Rays with Differentiable ROI Proposal Networks and Soft ROI Pooling}
    \author{
        Philip Müller, Felix Meissen, Georgios Kaissis and Daniel Rueckert, \IEEEmembership{Fellow, IEEE}
        \thanks{
        Philip Müller (philip.j.mueller@tum.de), Felix Meissen (felix.meissen@tum.de), Georgios Kaissis (g.kaissis@tum.de), and Daniel Rueckert (daniel.rueckert@tum.de) are  with the Chair for AI in Medicine and Healthcare (I31), School of Computation, Information and Technology, TU Munich, 85748 Garching
        }
        \thanks{Georgios Kaissis is with the group for Reliable AI, Institute for Machine Learning in Biomedical Imaging, Helmholtz Munich, Germany}
        \thanks{Daniel Rueckert is with the Biomedical Image Analysis Group, Imperial College London, London SW7 2AZ, UK}
        \thanks{Philip Müller and Felix Meissen contributed equally.}
    }
    \maketitle


    \input{sections/00_abstract}
    \input{sections/01_introduction}
    \input{sections/02_related_work}
    \input{sections/03_method}
    \input{sections/04_experiments}
    \input{sections/05_discussion}

    \input{nomenclature}
    \printnomenclature

    \bibliographystyle{IEEEtran.bst}
    \bibliography{ms}

\end{document}

%% file: sections/00_abstract.tex
\begin{abstract}
Weakly supervised object detection (WSup-OD) increases the usefulness and interpretability of image classification algorithms without requiring additional supervision.
The successes of multiple instance learning in this task for natural images, however, do not translate well to medical images due to the very different characteristics of their objects (i.e.\ pathologies).
In this work, we propose Weakly Supervised ROI Proposal Networks (WSRPN), a new method for generating bounding box proposals on the fly using a specialized region of interest-attention (ROI-attention) module.
WSRPN integrates well with classic backbone-head classification algorithms and is end-to-end trainable with only image-label supervision.
We experimentally demonstrate that our new method outperforms existing methods in the challenging task of disease localization in chest X-ray images.
Code: \url{https://github.com/philip-mueller/wsrpn}
\end{abstract}

%% file: sections/01_introduction.tex
\section{Introduction}\label{sec:introduction}

\IEEEPARstart{O}{bject} localization is a vital task in computer vision.
It is not only useful for many of the downstream tasks but is also a crucial factor for the interpretability of machine learning models.
However, especially in medical images, localization labels such as bounding boxes are costly and difficult to obtain as they require vast amounts of working hours from trained professionals.
Image labels, on the other hand, are easier to collect and can be mined from radiology reports associated with most existing medical images \cite{chexpert,cxr8}. This makes weakly supervised object detection (WSup-OD) a promising approach for the localization of diseases in medical images. It only requires image-level labels for training, allowing the use of such automatic collection approaches and thus making localization tractable for a wider range of medical applications.
WSup-OD has a long history in natural images \cite{is_localization_free,wsddn,contextlocnet}.
The SOTA methods here use multiple instance learning (MIL) \cite{MIL}, where bounding box proposals for each image are selected using algorithms such as Selective Search (SS) \cite{selective_search} or Edge Boxes (EB) \cite{edge_boxes}.
These algorithms, however, generate box proposals based on heuristics for objects in natural images and are not suited for detecting diseases in chest X-ray images, as the latter ones have very different characteristics and are more subtle.
Selective Search produces box proposals by over-segmenting the image based on pixel intensities.
Since pathologies in chest X-rays are not characterized by unique local intensities, the Selective Search algorithm is likely to not focus on them.
The Edge Boxes algorithm is based on the observation that in natural images, edges tend to correspond to object boundaries and, thus, searches for regions that wholly enclose edge contours.
This method again delivers unsatisfactory results for chest X-rays, as diseases here oftentimes do not have clear edges, and even existing boundaries are often not visible in summation images because they are covered by dense, radiopaque masses along the viewing direction.
That is why WSup-OD literature in medical images so far has mostly used CAM-based approaches \cite{cxr8,chexnet,stl,wsup_thoracic,wsup_covid} that extract boxes from heatmaps. However, these approaches are known to exhibit sub-par performance \cite{survey}.

\begin{figure}[t]
    \centering
    \includegraphics[width=\linewidth]{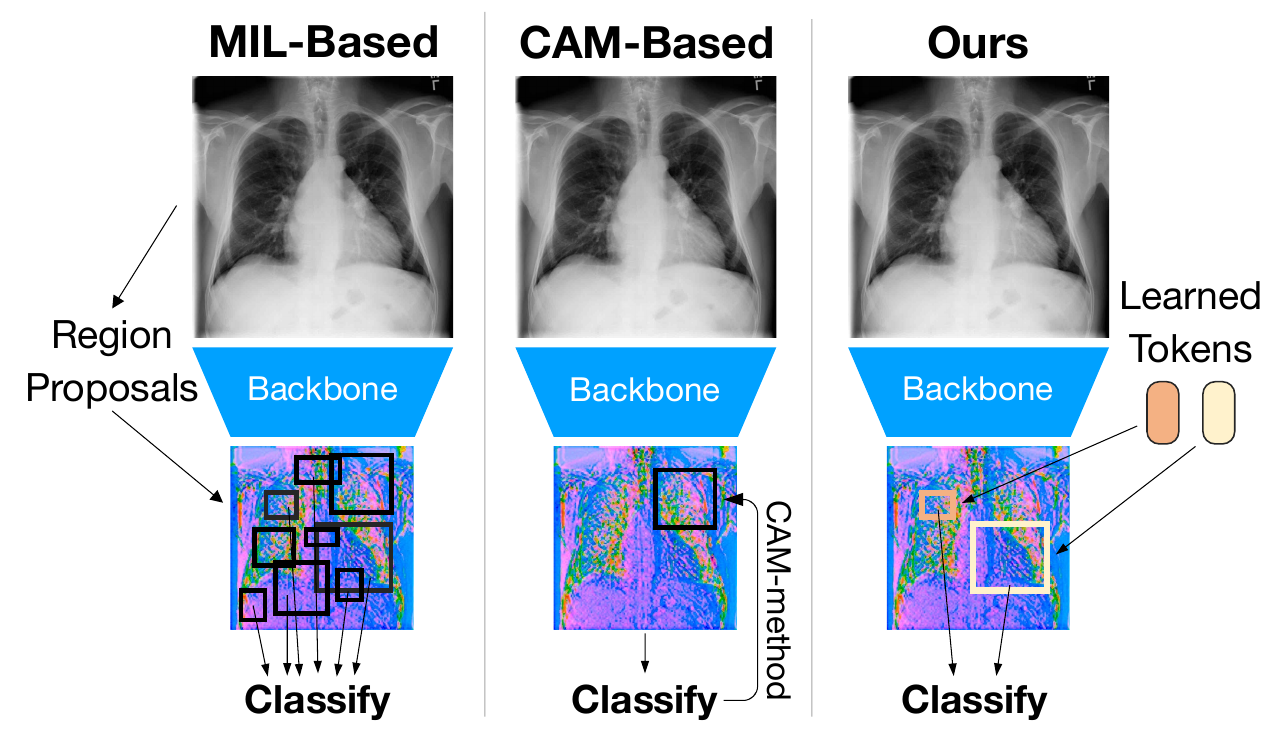}
    \caption{Schematic illustration of MIL-based, CAM-based, and our novel WSRPN approach.}
    \label{fig:compare_approaches}
\end{figure}

To address this issue, we propose Weakly Supervised ROI Proposal Networks (WSRPN), a novel paradigm for WSup-OD in medical images.
The bounding box proposals of our method are learned end-to-end and are predicted on the fly during the forward pass through an attention mechanism similar to DETR \cite{detr}.
To summarize, our contributions are the following:
\begin{itemize}
    \item We propose WSRPN, a novel, learnable, end-to-end trainable, and fully differentiable box-proposal algorithm for weakly supervised object detection in medical images.
    \item We set a new state-of-the-art for weakly supervised object detection on the challenging and commonly used CXR8 \cite{cxr8} dataset.
    \item To the best of our knowledge, we provide the first multiple-instance learning method successfully trained on this dataset.
\end{itemize}


%% file: sections/02_related_work.tex
\section{Related work}\label{sec:related_work}

\subsection{Weakly supervised object detection on natural images}
So far, most works in WSup-OD have focused on natural images in datasets such as PASCAL VOC \cite{pascal-voc-2007,pascal-voc-2012}, COCO \cite{cocodataset}, ILSVRC \cite{ilsvrc2016}, and CUB-200-2011 \cite{cub-200-2011}.
The two dominant approaches in the field are Multiple Instance Learning (MIL) and generating bounding boxes from Class Activation Maps (CAM). \cref{fig:compare_approaches} illustrates how these two approaches compare to our proposed method.

\paragraph{MIL}
In MIL-based approaches, each image is considered a bag of instances (regions). Every bag with a positive class label contains at least one positive region.
A MIL model is trained only with image labels by assigning every region the label of the whole bag.
After training with a large corpus of diverse images, the model becomes invariant to uncorrelated variations and gives higher scores to the most discriminative regions in an image.
To identify likely regions of objects in an image, region-proposal-algorithms, such as Selective Search \cite{selective_search} or Edge Boxes \cite{edge_boxes}, are commonly used \cite{wsddn,oicr,pcl,slv,contextlocnet}.
The seminal work here is by Bilen and Vedaldi \cite{wsddn}, who extract a feature vector for each region from a backbone network using a Spatial Pyramid Pooling (SPP) layer \cite{spp} and subsequently classify each region with a detection (\emph{is it an object?}) and a classification (\emph{which class?}) branch.
This method, however, tends to assign higher scores to the most discriminative regions in an image, which do not necessarily cover the whole extent of an object.
Subsequent work has, thus, mainly focused on solving the most discriminative region problem by refining the predictions iteratively using multiple refinement streams \cite{oicr}, 
incorporating the scores of larger context around the region \cite{contextlocnet},
clustering spatially adjacent regions of the same class \cite{pcl},
or maximizing the loss for the most discriminative region to force the model to focus on larger regions \cite{icmwsd}.
Very recently, Liao \textit{et al.} \cite{liao2022end} have proposed a novel method that uses Class Activation Maps as pseudo-ground-truth and cross-attention with learnable tokens to predict bounding boxes. Unlike our proposed WSRPN, however, their method is not fully differentiable, therefore limiting its use in more complex end-to-end models (c.f. \cref{sec:discussion}).

\paragraph{CAM}
The idea of using Class Activation Mapping for weakly-supervised object detection was first proposed by Zhou \textit{et al.} \cite{cam}. This method leverages the weights of the final classification layer to classify each patch in the un-pooled feature map and create an activation heatmap for each class that can be thresholded and used for object detection.
A similar idea was proposed by Pinheiro \textit{et al.} \cite{lse}.
However, they first classified each patch in the feature map and then aggregated the resulting scores via LSE pooling, alleviating the need to create heatmaps via CAM.
The authors of WELDON \cite{weldon} use max-min pooling instead to incorporate negative evidence in the final classification and, thus, create better class contrast between the regions.
Similar heatmaps are created via GradCAM \cite{gradcam}, which uses the gradients w.r.t. the feature map instead of the classifier weights.
Just like for MIL-based models, several approaches have been made to solve the most-discriminative-region problem for CAM-based models.
In ACoL, for example, Zhang \textit{et al.} \cite{acol} follow an idea similar to ICMWSD \cite{icmwsd}, masking out the most discriminative regions to make the model focus more on secondary features.

\subsection{Weakly supervised object detection in medical images}
WSup-OD is an underrepresented topic in the medical literature and is mainly focused on established CAM-based approaches from natural images. 
Along with the CXR8 dataset, Wang \textit{et al.} \cite{cxr8} proposed a model for WSup-OD. It uses CAMs for detection and LSE pooling \cite{lse} instead of average- or max-pooling.
The authors of CheXNet \cite{chexnet} also relied on the simple CAM approach for object localization in the chest X-ray images of the CXR8 dataset.
To guide the initially unstable localization in early epochs, Hwang and Kim \cite{stl} start with training for classification and gradually shift the focus towards detection using a dedicated branch for each of the two tasks. 
Their detection branch outputs a heatmap as in \cite{lse} to localize tuberculosis in chest X-ray images.
In \cite{wsup_thoracic}, the authors extended the work of Pinheiro \textit{et al.} \cite{lse} by using a multi-channel map for each class and employing max-min pooling as in \cite{weldon} to better localize diseases in CXR8.
Yu \textit{et al.} \cite{agxnet} included anatomical information from radiology reports to guide localization.
Lastly, Tang \textit{et al.} \cite{tang2018attention} improve the results of \cite{cxr8} on CXR8 by employing a curriculum learning strategy based on Disease Severity Labels mined from radiology reports and using attention guidance to improve localization performance.

However, none of the above works in the medical domain provides quantitative results of standard metrics in object detection, such as mean Average Precision, limiting the comparability and quantification of their localization performance.

%% file: sections/03_method.tex
\section{Method}\label{sec:method}

\begin{figure}[t]
    \centering
    \includegraphics[width=.45\textwidth]{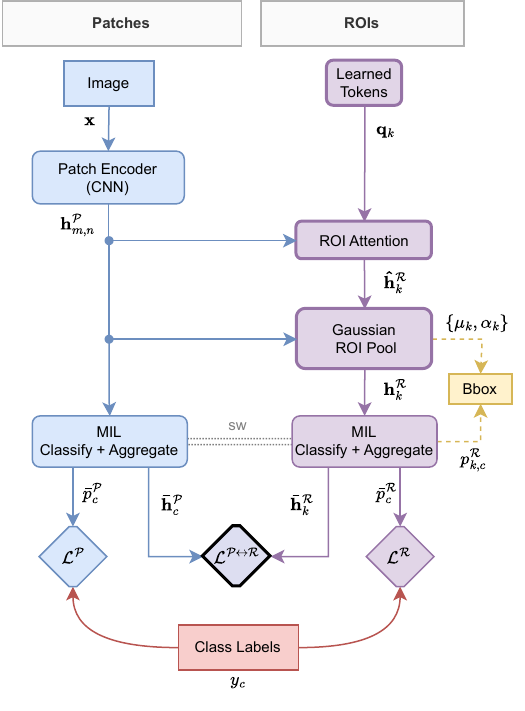}
    \caption{Overview of our model architecture. We show the patch branch (blue) and the ROI branch (purple), each with the encoding steps, MIL classification and aggregation, and the loss functions. Components typically used in a MIL model are colored in blue. Our key contributions are outlined with bold lines. \say{sw} stands for shared weights. Yellow denotes parts of the bounding box prediction.}
    \label{fig:model_overview}
\end{figure}

\subsection{Overview}
In our weakly supervised object detection setting, we assume that we are given an image that is labeled with a set $\mathcal{C}$ of non-exclusive classes, i.e.\, there is one binary classification label $y_{c} \in \{0, 1\}$ per class $c \in \mathcal{C}$ resulting in a multilabel binary classification task. Given only these per-image labels but without any bounding box supervision, we then learn an object detection model.

\cref{fig:model_overview} provides an overview of our method WSRPN.
It is based upon the MIL framework \cite{MIL,lse}, where \emph{regions-of-interest (ROIs)}, i.e.\, bounding boxes, are predicted using a bounding box proposal algorithm.
Following the findings of \cref{sec:introduction}, we however cannot use one of the classical, heuristic bounding box proposal algorithms but instead learn the algorithm end-to-end as a fully differentiable component of our network.
We, therefore, follow DETR\cite{detr} and use learned ROI query tokens attending to patch features (computed by a CNN backbone) and a box prediction network applied to the resulting ROI features.
However, since we do not have supervision for the box proposals, the DETR loss function cannot be applied.
To ensure that the predicted box parameters are meaningful (i.e.\, focus on relevant regions), we apply a Gaussian-based soft approximation of ROI pooling to aggregate ROI features from the patch features.
Using a Gaussian distribution during soft ROI pooling introduces an inductive bias that assures that ROI features represent locally restricted regions around the predicted center coordinates of the ROI.
The resulting ROI features are then classified and aggregated following the MIL framework, such that they can be trained using per-image class labels. 
Having only weak supervision, training the ROI proposals directly can lead to instabilities where the bad quality of box proposals during early training stages makes refining these proposals hard.
We thus propose a two-branch approach where in the first branch the MIL framework is applied to patches (we denote the patch branch by $\mathcal{P}$), while the second branch (denoted by $\mathcal{R}$) is designated to ROIs as described.
We train both branches using a loss per branch and also introduce a consistency loss, assuring that the ROI proposals are aligned with discriminative patches. 

In \cref{sec:patch_branch} and \cref{sec:roi_branch}, we describe the details of the patch and ROI branch, respectively, and in \cref{sec:loss_wsup}, we describe how these branches can be trained using weak supervision from classification labels.

\subsection{Patch branch}\label{sec:patch_branch}
\paragraph{Patch encoder}
In the patch branch, we first encode each image into $H \times W$ patches using the CNN backbone (we use DenseNet121\cite{densenet}).
These patches are then projected to the model dimension $d$, and 2D cosine position encodings~\cite{cosine_pos_2d,transformer} are added.
We denote the resulting embeddings of patch $(m, n)$ as $\hp \in \mathbb{R}^d$, where $m \in \{1, \dots, H\}$ is the $y$-index and $n \in \{1, \dots, W\}$ is the $x$-index of the patch.

\paragraph{Patch classification}
We now follow the MIL\cite{MIL,lse} approach and classify each patch $(m, n)$ into the classes in $\mathcal{C}$, but also predict an additional no-finding (i.e.\ background) class, denoted as $\nf$.
We compute the class logits $\tilde{p}^\mathcal{P}_{m, n, c}$ of all classes in $\mathcal{C}$ and the no-finding class $\nf$ by applying a multi-layer perceptron (MLP) to the corresponding patch features $\hp$ and then compute the class probabilities $\ppc$ via
\begin{align}\label{eq:prob_normalization}
\begin{split}
    \ppnf &= \phi\left(\tilde{p}^\mathcal{P}_{m, n, \nf}\right) \,,\\
    \ppc &= (1 - \ppnf)\cdot\phi\left(\tilde{p}^\mathcal{P}_{m, n, c}\right) \quad \forall c \in \mathcal{C}  \,,
\end{split}
\end{align}
where $\phi$ is the sigmoid function. Patches with large no-finding probabilities $\ppnf$ receive lower probabilities for other classes $c \neq \nf$.
Note that the other classes do not influence each other (i.e.\, each class is considered as a binary classification task) and are thus non-exclusive.
We found this approach more effective than having exclusive classes using Softmax.

\paragraph{Aggregation of patch probabilities}
Further following the MIL framework, we now obtain a single per-image probability for each class $c$ by aggregating the probabilities of all the patches using the \emph{LogSumExp (LSE)} function~\cite{lse} as a smooth approximation of max pooling as in\cite{lse,cxr8}, where we set the scaling hyperparameter $r$ to $5.0$.
We again assume multilabel binary classes, i.e.\, different classes $c$ are treated independently of each other instead of being exclusive. The aggregated probabilities $\aggppc$ of classes $c \in \mathcal{C}$ are thus computed as
\begin{align}
    \aggppc = \lse_{m, n}\left(\ppc\right) \qquad \forall c \in \mathcal{C} \,.
\end{align}
The no-finding class $\nf$ is considered a special case, and we aggregate it in two different ways: (i) following the \texttt{OR} logic, denoted by $\nfor$, where the class is considered positive if there is any positive patch (similar to the other classes),
and (ii) following the \texttt{AND} logic, denoted by $\nfand$, where it is considered positive only if all patches are positive, i.e.\, where there is no finding in the whole image. Case (ii) is implemented by inverting the probabilities of $\ppnf$ before LSE pooling.
The \texttt{OR} approach assures that there are always no-finding patches in an image, i.e.\, not all patches should be assigned a class, while the \texttt{AND} approach assures that in samples without any other classes, there are only no-finding patches.

\subsection{ROI branch}\label{sec:roi_branch}
\begin{figure}[t]
    \centering
    \includegraphics[trim=25 0 55 0, clip,width=.47\textwidth]{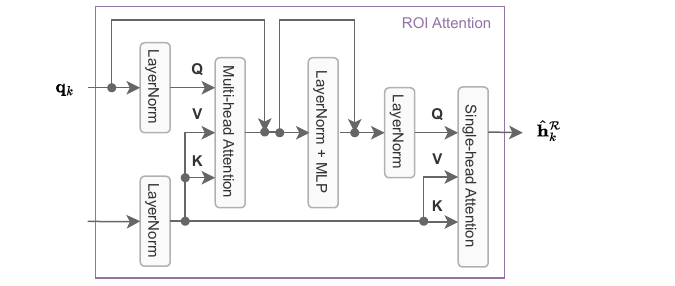}
    \caption{ROI attention component from our ROI branch. Using cross-attention, ROI tokens $\{\bm{q}_k\}$ gather relevant information from the patch features $\{\hp\}$ to compute the ROI features $\{\hat{\bm{h}}^\mathcal{R}_{k}\}$.}
    \label{fig:roi_attention}
\end{figure}

\paragraph{ROI attention}
In the ROI branch, we use $K$ learned ROI tokens $\bm{q}_k$ (where $K$ is a hyperparameter, set to 10 in our experiments).
Given a ROI token $\bm{q}_k$, we now use our \emph{ROI attention} component to gather relevant information from the patch features $\hp$ to compute the ROI features $\hat{\bm{h}}^\mathcal{R}_{k}$.
As shown in \cref{fig:roi_attention}, the ROI attention component first performs multi-head cross attention\cite{transformer} with ROI tokens used as queries and patch features used as keys and values.
It then further processes the resulting token features using an MLP and a single-head cross-attention layer, where patch features are again used for keys and values. 

\paragraph{Box prediction and Gaussian ROI pooling}

Given the token features $\hat{\bm{h}}^\mathcal{R}_{k}$ of token $k$, 
we now predict its box center coordinates $\bm{\mu}_{k}$ and size $\bm{\sigma}_{k}$, each relative to the image size.
We assume that relevant features within each ROI are roughly distributed following a normal distribution around the box center.
Following this assumption, we now propose a smooth, and therefore differentiable, approximation of (hard) ROI pooling \cite{fast_rcnn}.
For each ROI $k$, we compute a soft receptive field (i.e.\ attention map) $A_{k, m, n}$ over all patches $(m, n)$, centered over the ROI center $\bm{\mu}_{k}$ and with its scale (i.e.\ width and height) controlled by $\bm{\sigma}_{k}$.
We compute the receptive field $A_{k, m, n}$, which is proportional to the probability density function of a 2D multivariate Gaussian with independent $x$ and $y$ components (i.e. with zero covariance), as
\begin{align}
\begin{split}
    A_{k, m, n} \propto & \exp \left[ {-\frac{1}{2} \left( \frac{\frac{m+0.5}{H} - \mu_{k, y}}{\sigma_{k, y}} \right)^2 } \right] \times \\
    & \exp \left[ {-\frac{1}{2} \left( \frac{\frac{n+0.5}{W} - \mu_{k, x}}{\sigma_{k, x}} \right)^2 } \right] \,.
\end{split}
\end{align}
Examples of such receptive fields $A_{k, m, n}$ are shown in \cref{fig:results_with_gaussians}.
Finally, we aggregate the patch features $\hp$ for each ROI $k$ using the receptive field $A_{k, m, n}$ to get the final ROI features $\hr$. 

\paragraph{ROI classification}
We assign each ROI $k$ a probability $\prc$ for each $c \in \mathcal{C}\cup\{\nf\}$ by using the classifier from the patch branch, including sharing the same weights, and applying it to the ROI features $\hr$. 

\paragraph{MIL aggregation of ROI probabilities}
As in the patch branch, we again follow the MIL framework to aggregate the ROI probabilities $\prc$ over the whole image.
However, instead of using LSE, we found the $\noisyor$~\cite{noisyOR1,noisyOR2,noisyOR3} aggregation strategy more effective. $\noisyor$ and its counterpart $\noisyand$ are defined as follows:
\begin{align}\label{eq:noisy_or}
    \noisyor_{k}\left(\bm{p}\right) &= 1 - \prod_k \left(1 - p_k\right) \,,\\
    \noisyand_{k}\left(\bm{p}\right) &= \prod_k p_k \,.
\end{align}
Using these aggregation functions, we now compute the aggregated ROI probabilities $\aggprc$ for $c \in \mathcal{C}$:
\begin{align}\label{eq:noisy_and}
    \aggprc &= \noisyor_{k}\left(\prc\right) \qquad \forall c \in \mathcal{C} \,.
\end{align}
We again consider the special cases for the no-finding class and aggregate with the \texttt{OR} ($\aggprnfor$) and \texttt{AND} ($\aggprnfand$) logic.

\subsection{Weakly supervised loss function}\label{sec:loss_wsup}
Our weakly supervised loss function is defined as
\begin{align}
    \mathcal{L} = \lossp{} + \lossr{} + \mathcal{L}^{\mathcal{P}\leftrightarrow\mathcal{R}} \,,
\end{align}
where $\lossp{}$ trains the patch branch, $\lossr{}$ trains the ROI branch, and $\mathcal{L}^{\mathcal{P}\leftrightarrow\mathcal{R}}$ assures that both branches are mutually consistent. 
The branch-specific loss functions ($\lossp{}$ and $\lossr{}$) each consist of two components: (i) a multilabel binary cross entropy loss $\loss{bce}$ applied on aggregated patch or ROI probabilities for providing strong gradients, and (ii) a supervised contrastive loss $\loss{supcon}$ \cite{supcon} applied on per-class features from the patch or ROI branch for pushing the patches and ROIs to focus on discriminative regions. We therefore define the branch-specific loss functions as
\begin{align}
    \lossp{} = \lossp{bce} + \lossp{supcon}\,, \qquad  \lossr{} = \lossr{bce} + \lossr{supcon}\,.
\end{align}

\paragraph{Multilabel binary cross entropy}
For the multilabel binary cross entropy losses $\lossp{bce}$ and $\lossr{bce}$, we use the per-image binary labels $\yc$ with $c\in\mathcal{C}$ and $\yc \in \{0, 1\}$.
We additionally define the no-finding label with AND logic $\yc[\nfand]$ as true only if no other classes are true, i.e. $\yc[\nfand] = 1 - \max_{c \in \mathcal{C}} \yc$, as then all patches/ROIs should be classified as no-finding, and the no-finding label with OR logic $\yc[\nfor]$ as always true, i.e. $\yc[\nfor] = 1$, as there should always be some patch/ROI that contains no finding.
The losses $\lossp{bce}$ and $\lossr{bce}$ are weighted multilabel binary cross entropy losses over the classes $\mathcal{C}\cup\{\nfand,\nfor\}$
and are applied to the aggregated patch probabilities $\aggppc$ and ROI probabilities $\aggprc$, respectively.

\paragraph{Supervised contrastive loss}
The losses $\lossp{supcon}$ and $\lossr{supcon}$ are based on the supervised contrastive loss~\cite{supcon}, which is an NTXent-based loss function~\cite{SimCLR} where positive pairs are defined based on label supervision.
We consider each class $c \in \mathcal{C}$ (but not the no-finding class $\nf$) independently (as a binary label) and define the set of positive samples $j$ for each sample $i$ and class $c$ as
\begin{align}
    P(i, c) = \left\{j \in \{1, \dots, N\}: y^{(i)}_{c} = y^{(j)}_{c} \right\} \,.
\end{align}

Following this setting, we require (sample-wise) per-class features for each $c \in \mathcal{C}$, which are computed once from the patch branch (for $\lossp{supcon}$) and once from the ROI branch (for $\lossr{supcon}$), and are denoted by $\agghpc \in \mathbb{R}^d$ and $\agghrc \in \mathbb{R}^d$, respectively.
We consider the class probabilities of each patch ($\ppc$) or ROI ($\prc$) and compute the per-class features $\agghpc$ and $\agghrc$ as a weighted sum of all patch ($\hp$) and ROI ($\hr$) features, respectively, with weights computed as their (normalized) class probabilities.
Finally, we project the results using an MLP. 
Similarly, we compute $\agghrc$ from ROI features $\hr$ considering $\prc$, where the MLP is shared between both branches.

Given these aggregated patch features $\agghpc$ and ROI features $\agghrc$, respectively, as representations for class $c$ in sample $i$, the losses $\lossp{supcon}$ and $\lossr{supcon}$ follow the following form:
\begin{align}
    \begin{split}
    \loss{supcon} =& \frac{1}{N\lvert\mathcal{C}\rvert}\sum_{i=1}^N\sum_{c \in \mathcal{C}}\frac{1}{\lvert P(i, c)\rvert} \sum_{j\in P(i, c)}  \\
    & \qquad \log \frac{e^{\cos\left(\bar{\bm{h}}^{(i)}_{c}, \bar{\bm{h}}^{(j)}_{c}\right) / \tau}}{\sum_{j'=1}^N e^{\cos\left(\bar{\bm{h}}^{(i)}_{c}, \bar{\bm{h}}^{(j')}_{c}\right) / \tau}} \,.
    \end{split}
\end{align}

\paragraph{Patch-ROI consistency regularizer}
To stabilize training and guide the generation of useful features in the ROI branch, we introduce a consistency regularization loss $\losspr$. This loss ensures the agreement between the spatial distribution of class features of the ROI- and the patch branch.
To calculate this agreement, we first need to compute spatial class-distribution probabilities (i.e.\ patch-wise class probabilities) for both the patch- and the ROI branch. 
While these class probability maps already exist for the patch branch (cf.\ $\ppc$ from \cref{eq:prob_normalization}), getting them for the ROI branch requires further steps:

For the ROI branch, we know the class probabilities $\prc$ and the spatial distribution (given by the soft receptive field $A_{k, m, n}$) of each ROI. We use these to compute the spatial class map $p^{\mathcal{R} \rightarrow \mathcal{P}}_{m, n, c}$ of each class $c$ for each patch $(m, n)$ as follows:
\begin{align}
    p^{\mathcal{R} \rightarrow \mathcal{P}}_{m, n, c} &= 
    \noisyor_k \left(A_{k, m, n} \cdot \prc \right) \qquad \forall c \in \mathcal{C} \,.
\end{align}
For the no-finding class $\nf$, we consider the assigned patches of ROIs with high no-finding probabilities $\prnf$ as well as patches where ROIs have low attention $A_{k, m, n}$ and use $\noisyand$ pooling over the ROIs:
\begin{align}
    p^{\mathcal{R} \rightarrow \mathcal{P}}_{m, n, \nf} &= 
    \noisyand_k \left(A_{k, m, n} \cdot \prnf + (1- A_{k, m, n})\right)
\end{align}
This assures that patches that are only marginally considered during Gaussian ROI pooling but have high probabilities in real classes $c \in \mathcal{C}$, receive high probabilities for the no-finding class.

We now define the consistency loss $\losspr$ using the empirical KL-divergence $D_\text{KL}$ from the newly computed spatial class map $p^{\mathcal{R} \rightarrow \mathcal{P}}_{m, n, c}$ (from the ROI branch) to the (original) spatial class map $\ppc$ (from the patch branch):
\begin{align}
    \losspr &= \frac{1}{HW}\sum_{m, n}^{H, W}
    \kl_{c \in \mathcal{C} \cup \{\nf\}} \left[p^{\mathcal{P}}_{m, n, c} \Bigg\| p^{\mathcal{R} \rightarrow \mathcal{P}}_{m, n, c} \right]
\end{align}

\subsection{Inference}
During inference, we initially predict one box for each ROI $k$.
Center position $\bm{\mu}_{k}$ and box size $\bm{\sigma}_{k}$, computed during box prediction, are used as box parameters. 
We compute the predicted class $c^*_{k} \in \mathcal{C}$ of ROI $k$ as $c^*_{k} = \argmax_{c \in \mathcal{C}} \prc$ and use $\prc[c^*_{k}]$ as its confidence score.
Finally, we apply standard post-processing as it will be described in \cref{sec:experiments}.

%% file: sections/04_experiments.tex
\section{Experiments}\label{sec:experiments}
We show the effectiveness of our Weakly Supervised ROI Proposal Network on the task of disease localization in chest X-ray images.

\paragraph{Dataset and evaluation metrics}
We follow previous works \cite{cxr8,chexnet,wsup_thoracic} and evaluate on the challenging ChestXray-8 (CXR8) dataset \cite{cxr8}.
The dataset consists of 108\,948 X-ray images from the National Institutes of Health Clinical Center in the US.
The dataset contains labels for eight different disease types and \say{no-finding} ($\nf$).
Each image can have more than one positive label, turning the task into a multi-class classification problem.
All labels were automatically mined from associated radiology reports with an algorithm that achieved an F1 score of 0.90 on an external dataset. The labels, thus, include a significant amount of noise, making the dataset challenging, even for classification.
Additionally, the dataset contains 984 bounding boxes on 882 images from unique patients, hand-labeled by a board-certified radiologist.
From the images with bounding boxes, we used 50\% for validation and kept the other 50\% as a held-out test set.
The images of patients that were not included in the validation or test sets were used for training.

To compare the performance of our proposed model with the baselines, we report the Robust Detection Outcome (RoDeO) \cite{rodeo}, a recently proposed metric for object detection in medical images, such as Chest X-rays, that reflects the clinical requirements for object detection methods better than other metrics and further gives insights about strengths and weaknesses of the models.
Additionally, we report standard metrics, such as Average Precision (AP) and localization accuracy (loc-acc) at two different IoU thresholds ($0.3$ and $0.5$).
Note, however, that loc-acc is biased to favor models that predict fewer boxes.

\begin{table*}[ht!]
    \centering
    \caption{Mean and standard deviation of our method WSRPN against baseline methods on RoDeO, AP, and localization accuracy. The best method per metric is marked in bold. Our method outperforms all baselines on all object detection metrics, setting a new state-of-the-art for weakly supervised object detection on the challenging CXR8~\cite{cxr8} dataset.}
    \label{tab:results_thres}
    \begin{tabular}{@{}lcccccccc@{}}
        \toprule
        \multirow{2}{*}{Method}
        & \multicolumn{4}{c}{\textbf{RoDeO} [\%]} & \multicolumn{2}{c}{AP [\%]} & \multicolumn{2}{c@{}}{loc-acc}  \\
        \cmidrule(lr){2-5} \cmidrule(lr){6-7} \cmidrule(lr){8-9} & cls & loc & shape & total & IoU@0.3 & IoU@0.5 & IoU@0.3 & IoU@0.5 \\
        \midrule
        \rowcolor{Gray}WSRPN (ours) & \textbf{31.7\std{2.4}} & \textbf{44.1\std{1.6}} & \textbf{29.4\std{1.0}} & \textbf{34.0\std{1.3}} & \textbf{9.44\std{0.90}} & \textbf{6.34\std{0.86}} & \textbf{0.78\std{0.00}} &  \textbf{0.77\std{0.00}}  \\ 
        CheXNet \cite{chexnet} & 19.8\std{1.1} & 19.9\std{0.7} & 10.9\std{0.4} & 15.6\std{0.5} & 8.26\std{0.81} & 3.32\std{0.56} & 0.55\std{0.00} & 0.52\std{0.00} \\
        \rotatebox[origin=c]{180}{$\Lsh$} w/ noisyOR aggregation & 23.9\std{1.1} & 20.1\std{0.7} & 12.2\std{0.4} & 17.3\std{0.5} & 8.45\std{0.92} & 1.13\std{0.33} & 0.59\std{0.00} & 0.55\std{0.00} \\  
        \rotatebox[origin=c]{180}{$\Lsh$} w/ $\lossp{supcon}$ & 20.8\std{1.2} & 22.3\std{0.8} & 11.9\std{0.5} & 17.0\std{0.6} & 7.44\std{0.83} & 4.00\std{0.65} & 0.58\std{0.00} & 0.56\std{0.00} \\ 
        STL \cite{stl} & 19.0\std{1.0} & 18.5\std{0.6} & 10.6\std{0.4} & 14.9\std{0.5} & 8.59\std{0.78} & 2.73\std{0.58} & 0.54\std{0.00} & 0.50\std{0.00} \\
        GradCAM \cite{gradcam} & 17.6\std{1.2} & 17.5\std{0.6} & 9.8\std{0.4} & 13.9\std{0.5} & 7.07\std{0.87} & 0.18\std{0.14} & 0.54\std{0.00} & 0.51\std{0.00} \\ 
        CXR \cite{cxr8} & 19.9\std{1.1} & 19.5\std{0.7} & 11.3\std{0.4} & 15.8\std{0.5} & 8.54\std{0.87} & 1.46\std{0.41} & 0.55\std{0.00} & 0.51\std{0.00} \\ 
        WELDON \cite{weldon} & 18.5\std{1.3} & 20.6\std{0.7} & 12.1\std{0.4} & 16.2\std{0.5} & 6.76\std{0.82} & 0.48\std{0.27} & 0.56\std{0.00} & 0.52\std{0.00} \\ 
        MultiMap Model \cite{wsup_thoracic} & 21.0\std{1.2} & 20.0\std{0.7} & 11.6\std{0.4} & 16.3\std{0.5} & 7.53\std{0.80} & 1.53\std{0.41} & 0.57\std{0.00} & 0.53\std{0.00} \\ 
        LSE Model \cite{lse} & 20.0\std{1.1} & 21.3\std{0.7} & 11.5\std{0.4} & 16.3\std{0.5} & 3.07\std{0.60} & 0.58\std{0.29} & 0.56\std{0.00} & 0.54\std{0.00} \\ 
        ACoL \cite{acol} & 14.8\std{1.0} & 11.9\std{0.5} & 10.2\std{0.4} & 12.0\std{0.4} & 4.27\std{0.66} & 2.84\std{0.58} & 0.48\std{0.00} & 0.48\std{0.00} \\ 
        \bottomrule
    \end{tabular}
\end{table*}

\paragraph{Implementation details}
As the backbone for our model and all baselines, we used a Densenet121 \cite{densenet} as in \cite{chexnet,wsup_thoracic}, pre-trained on ImageNet \cite{imagenet}.
For all CAM-based methods that produce heatmaps, we adopted the bounding box generation method and parameters of Wang \textit{et al.} \cite{cxr8}, where the heatmaps are binarized, and box proposals are drawn around each connected component.
We extended this method to also produce class probabilities and confidence scores per box (c.f. supplementary material).
This is necessary to apply score-based postprocessing and compute the Average Precision metrics.
Unless indicated otherwise, we post-process the predictions of all models by keeping only the most confident predicted box per class (top1-per-class), a valid assumption in the CXR8 dataset that has at maximum one box for any class per image.
We implemented all models in PyTorch \cite{pytorch} and optimized them using AdamW \cite{adamw} with a learning rate of $1.5 \cdot 10^{-4}$, weight decay of $10^{-6}$, and gradient clipping at norm $1.0$.
All models were trained for a maximum number of 50000 iterations with early stopping (patience set to 10000) and a batch size of 128.
Finally, the checkpoint with the highest mAP on the validation set was chosen.
The images were resized to $224 \times 224$ pixels and normalized with the mean and standard deviation of the training dataset.
During training, we augmented the data by applying random color jitter and random Gaussian blurring, each with a probability of 50\%, using the Albumentations library \cite{albumentations}.
We applied two different random augmentations to each image to guarantee always at least one positive sample for the $\mathcal{L}_\text{supcon}$.
During validation or testing, no data augmentation was applied.
All of our experiments were performed on a single Nvidia RTX A6000 GPU. Our model trained for roughly 8 hours, requiring about 11 GB of GPU memory.

\subsection{Comparison with the baselines}
\cref{tab:results_thres} shows the results of our method and the baselines.
WSRPN significantly outperforms all weakly supervised baselines on all metrics by a large margin (Welch’s t-test, $p<0.001$).
Compared to the best baseline CheXNet w/ noisyOR aggregation, WSRPN achieves a relative improvement of $96.5\%$ in RoDeO score, setting a new state-of-the-art.
Especially, the box quality of our method is better than the baselines.
For the submetrics $\text{RoDeO}_{\text{loc}}$ and $\text{RoDeO}_{\text{shape}}$, the relative improvements to CheXNet w/ noisyOR aggregation are $119.4\%$ and $141.0\%$, respectively.
Also, in terms of AP and loc-acc, our method outperforms the baselines by a large margin, especially when more accurate localization is required.
At the IoU-threshold of $0.5$, we notice a relative improvement of $58.5\%$ in AP.
On the test set, WSRPN predicts, on average, $1.049$ boxes per sample, which much more closely resembles the $1.098$ true boxes per sample than CheXNet with $3.411$ (even after applying top1-per-class filtering).
This quality is also expressed by the much better loc-acc across all thresholds compared to the baselines.

All baselines in \cref{tab:results_thres} are CAM-based since MIL-based training (WSDDN \cite{wsddn}) did not converge on this challenging dataset and was thus excluded from the table.
A likely reason for the failure of WSDDN are the box proposal algorithms available for this method (SS and EB).
The box proposals of these algorithms have a significantly lower overlap with the objects in chest X-ray images than with those of natural images (c.f. \cref{tab:box_proposals_cxr_pascal}).
This strongly limits the detection performance of models building upon these proposals.

\begin{table}[htb]
    \centering
    \caption{We computed the average IoU of the target boxes in CXR8 \cite{cxr8} and PASCAL VOC 2007 \cite{pascal-voc-2007} and the boxes produced by the Selective Search (SS) \cite{selective_search} and Edge Boxes (EB) \cite{edge_boxes} algorithms. We only considered the predicted box with the highest IoU for every target box, making these numbers an upper bound for methods using SS or EB.}
    \label{tab:box_proposals_cxr_pascal}
    \begin{tabular}{lcc}
        \toprule
        Algorithm & CXR8 & PASCAL VOC 2007 \\
        \midrule
        Selective Search (fast) & 0.31 & 0.76 \\
        Selective Search (quality) & 0.37 & 0.82 \\
        Edge Boxes & 0.50 & 0.75 \\
        \bottomrule
    \end{tabular}
\end{table}

\begin{table*}[t]
    \centering
    \caption{Results per pathology of our model WSRPN and the best baselines. 
    Overall, our method WSRPN performs significantly better on five pathologies (atelectasis, cardiomegaly, effusion, mass, and nodule). On pneumothorax, it is competitive with the baselines, while on two pathologies (infiltration and pneumonia), it performs notably worse. However, WSRPN outperforms the baselines on all pathologies when considering localization and shape similarity.}\label{tab:patho_results}
    \begin{tabular}{@{}lcccccccc@{}}
    \toprule
    & \multicolumn{2}{c}{RoDeO cls [\%]} & \multicolumn{2}{c}{RoDeO loc [\%]} & \multicolumn{2}{c}{RoDeO shape [\%]} &\multicolumn{2}{c}{RoDeO total [\%]} \\
    \cmidrule(lr){2-3}\cmidrule(lr){4-5}\cmidrule(lr){6-7}\cmidrule(lr){8-9} Pathology & WSRPN & CheXNet noisyOR & WSRPN & CheXNet noisyOR & WSRPN & CheXNet noisyOR & WSRPN & CheXNet noisyOR \\
    \midrule
    Atelectasis  & \textbf{28.4\std{5.3}} & 24.2\std{3.0} & \textbf{34.7\std{3.5}} & 17.1\std{2.1} & \textbf{18.7\std{1.9}} & 7.8\std{1.0} & \textbf{25.3\std{2.3}} & 13.2\std{1.6} \\ 
    Cardiomegaly & \textbf{73.9\std{5.7}} & 46.9\std{5.5} & \textbf{95.0\std{1.4}} & 65.0\std{4.1} & \textbf{66.3\std{1.5}} & 38.0\std{2.4} & \textbf{76.5\std{2.5}} & 47.4\std{3.3} \\ 
    Effusion & \textbf{53.6\std{6.6}} & 26.6\std{2.7} & \textbf{34.0\std{3.6}} & 11.2\std{1.5} & \textbf{26.4\std{2.3}} & 9.9\std{1.1} & \textbf{34.8\std{3.0}} & 13.1\std{1.5} \\ 
    Infiltration & 2.2\std{3.0} & \textbf{16.7\std{2.6}} & \textbf{50.4\std{5.0}} & 15.2\std{1.9} & \textbf{30.6\std{2.8}} & 8.8\std{1.1} & 4.9\std{6.3} & \textbf{12.5\std{1.6}} \\ 
    Mass & \textbf{43.6\std{8.7}} & 17.7\std{3.4} & \textbf{36.3\std{5.5}} & 12.1\std{2.4} & \textbf{20.5\std{3.0}} & 5.4\std{1.2} & \textbf{29.9\std{3.9}} & 9.2\std{1.8} \\ 
    Nodule & \textbf{39.5\std{8.6}} & 19.9\std{3.9} & \textbf{6.2\std{2.6}} & 3.6\std{1.3} & \textbf{3.7\std{0.4}} & 1.4\std{0.2} & \textbf{6.3\std{1.3}} & 2.8\std{0.6} \\ 
    Pneumonia & 0.0\std{0.0} & \textbf{8.5\std{3.6}} & \textbf{57.7\std{4.3}} & 45.0\std{4.3} & \textbf{34.6\std{2.4}} & 24.7\std{2.5} & 0.0\std{0.0} & \textbf{16.0\std{5.0}} \\ 
    Pneumothorax & 22.6\std{7.2} & \textbf{37.5\std{5.0}} & \textbf{14.1\std{3.4}} & 13.8\std{2.4} & \textbf{20.2\std{2.6}} & 15.6\std{2.1} & 17.6\std{3.1} & \textbf{18.3\std{2.5}} \\ 
    \bottomrule
    \end{tabular}
\end{table*}

\subsection{Performance on different pathologies}

\begin{figure}[ht!]
    \centering
    \includegraphics[width=\columnwidth]{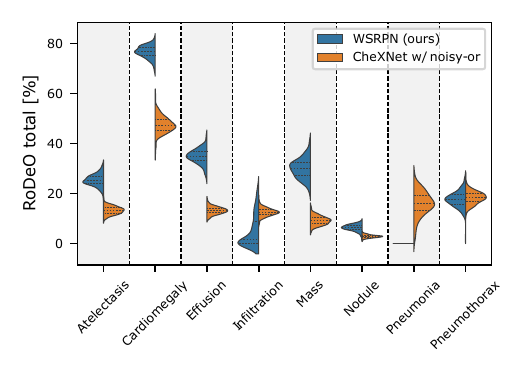}
    \caption{Comparison of the results per pathology between our method WSRPN and the best baseline on the bootstrapped ($N=250$) test set. On five pathologies (atelectasis, cardiomegaly, effusion, mass, and nodule), our WSRPN method performs significantly better, on pneumothorax, it is competitive with the baselines, while on two pathologies (infiltration and pneumonia), it performs worse.}
    \label{fig:violin}
\end{figure}
\begin{figure}[ht]
\centering
    \includegraphics[width=\columnwidth]{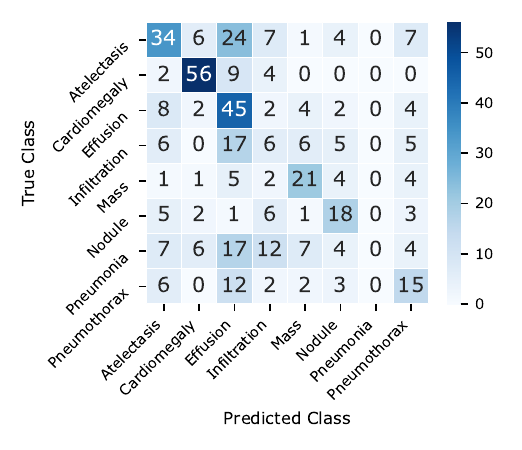}
    \caption{Confusion matrix for our proposed WSRPN. The matrix was generated from the 1-to-1 correspondences between predicted and ground-truth boxes after the matching step in RoDeO \cite{rodeo}.}
    \label{fig:confusion_matrix}
\end{figure}
\begin{figure*}[ht]
\centering
    \includegraphics[width=.98\textwidth]{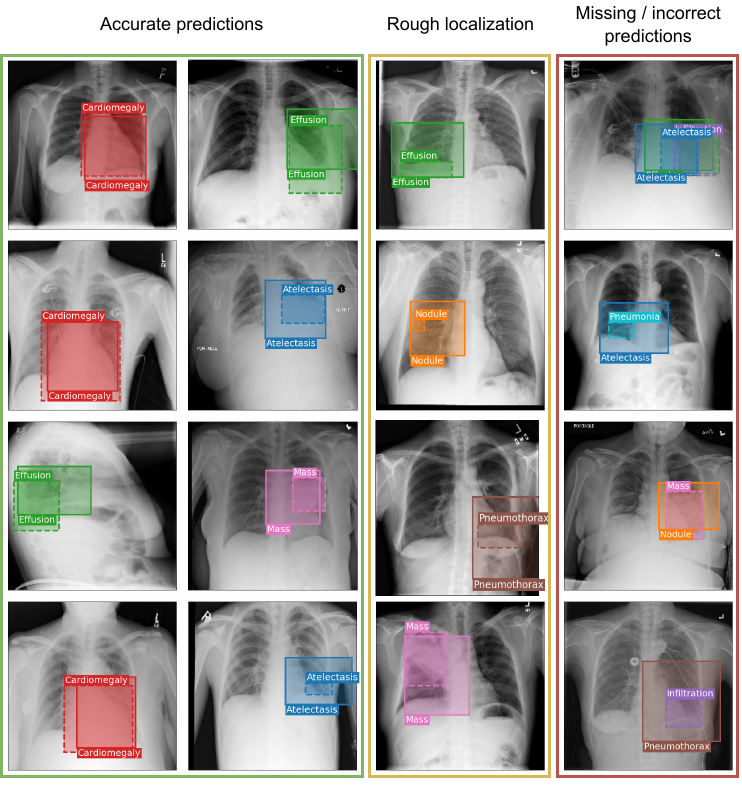}
    \caption{Qualitative results of some exemplary images. Left: successfully detected pathologies. Middle: Roughly localized correct predictions. Right: failure cases. Solid boxes are predictions. Dashed boxes are human-annotated targets.}
    \label{fig:results_with_gaussians}
\end{figure*}

In \cref{fig:violin} and \cref{tab:patho_results}, we study the results individually for each of the eight pathologies (atelectasis, cardiomegaly, effusion, infiltration, mass, nodule, pneumonia, pneumothorax) of the bootstrapped ($N=250$) test dataset. We compare our model WSRPN with the best baseline (CheXNet with $\noisyor$-aggregation). 
We observe (cf.\ \cref{fig:violin}) that on five pathologies (atelectasis,
cardiomegaly, effusion, mass, and nodule), our method WSRPN
performs significantly better than the baseline, often by large margins. On pneumothorax, it is competitive with the baseline, while on two pathologies (infiltration and pneumonia), it performs notably worse.
\cref{tab:patho_results} provides further explanations by distinguishing between the quality of classification, localization, and (box) shape similarity. We observe that the localization and shape quality of our model WSRPN outperforms the baseline for all eight pathologies by large margins. For localization, we observe relative improvements of $103\%$ for atelectasis, $46\%$ for cardiomegaly, $204\%$ for effusion, $232\%$ for infiltration, $200\%$ for mass, $72\%$ for nodule, $28\%$ for pneumonia, $2\%$ for pneumothorax, 
For shape similarity, these improvements are similarly significant.

We further show the confusion matrix for our proposed WSRPN in \cref{fig:confusion_matrix}.
Since confusion matrices are not trivial to generate for object detection problems, we computed them from the 1-to-1 correspondences between predicted and ground-truth boxes after the matching step in RoDeO \cite{rodeo}.
The figure confirms that the model often confuses infiltration with pneumonia and that it seems to fail to predict cases of pneumonia.
Both classes, however, are not well defined in CXR images.
Infiltration is an imprecise descriptive term used for accumulations of an abnormal substance in the lung, while pneumonia is a clinical diagnosis that can not solely be made from an X-ray image.
Pneumonia is further often detected by symptoms such as infiltrations and related to pleural effusions, in part explaining the confusions in the figure.

From these observations, we conclude the following: 
i) Our model WSRPN performs exceptionally well at localizing pathologies, while its classification capabilities reveal limitations on some classes. This can especially be observed for pneumonia and infiltration, where no or only a few bounding boxes are correctly classified, and on pneumothorax, where the baseline performs particularly well at classification. 
ii) Good localization and classification capabilities do not necessarily correlate between pathologies. For example, pneumonia is localized well but classified incorrectly, while nodules are classified quite well but are not located well.

\subsection{Qualitative results and failure cases}
\cref{fig:results_with_gaussians} shows example predictions of our model.
The first two columns show correctly detected pathologies.
The quality of these samples reflects the performance of our model for each class (c.f. \cref{tab:patho_results}): Cardiomegaly is detected nearly perfectly, but also effusion, atelectasis, and mass are often successfully detected by the proposed WSRPN.
Besides the successful cases, we mainly identified two types of failure cases, namely (i) imprecise prediction of the exact extent of the pathology and (ii) miss-classification or partial detection.

Examples of failure type (i) are shown in \cref{fig:results_with_gaussians}, column three (yellow column). Here, pathologies are detected and roughly localized. However, the predicted bounding boxes do not match the target boxes because their aspect ratios differ or the predicted box is too small or too large.
Such cases may be hard to tackle and may require semi-supervision, especially as the exact extent of pathologies can be hard to define and is often subjective. 
Generally, our model tends to produce larger boxes, which is especially problematic for classes with small boxes, such as nodules (c.f. \cref{tab:patho_results}).
However, the low performance of the baseline models indicates that this is a common problem of WSup-OD models.

Failure cases of type (ii) are shown in \cref{fig:results_with_gaussians}, column four (red column) and include cases where bounding boxes are predicted approximately correctly but with incorrect classes, especially if classes have similar clinical meaning (e.g.\, mass and nodule, row 3, col 4) or are correlated (e.g.\, pneumonia increasing the likelihood of atelectasis, row 2, col 4). In other such cases, classes are not detected at all, especially in samples with multiple overlapping boxes (row 1, col 4). We assume a significant part of this category of failure cases can be tackled by improving the model's classification performance on the dataset.

\subsection{Ablation studies}

\begin{table}[ht]
    \centering
    \caption{Ablation study on the loss function. We experimented with different combinations of the individual loss components. Our default configuration (using all components) is highlighted in grey.}
    \label{tab:ablation_loss}
    \renewcommand{\arraystretch}{1.2}
    \setlength{\tabcolsep}{2.2pt}
    \begin{tabular}{l|cc|cc|c|cc}
        \toprule
        Method  & $\lossp{bce}$ & $\lossp{supcon}$ & $\lossr{bce}$ & $\lossr{supcon}$ & $\losspr$ & RoDeO & AP@0.3 \\
        \midrule
        \rowcolor{Gray} WSRPN & \cmark & \cmark & \cmark & \cmark & \cmark & \textbf{34.0\std{1.3}} & \textbf{9.44\std{0.90} } \\ 
        \midrule
        no $\lossp{}$ &  &  & \cmark & \cmark & \cmark & 12.0\std{0.5} & 0.34\std{0.19} \\ 
        no $\lossr{}$ & \cmark & \cmark &  &  & \cmark & 18.7\std{0.7}& 9.00\std{0.94} \\ 
        no $\losspr$ & \cmark & \cmark & \cmark & \cmark & &18.0\std{0.8}  & 2.74\std{0.56} \\ 
        \midrule 
        no $\loss{bce}$ &  & \cmark &  & \cmark & \cmark & 6.3\std{0.4} & 0.16\std{0.12}\\ 
        no $\loss{supcon}$ & \cmark &  & \cmark &  & \cmark & 23.2\std{0.9} & 6.31\std{0.83} \\ 
        only $\loss{bce}$ & \cmark &  & \cmark &  & & 14.2\std{0.6} & 0.59\std{0.23} \\ 
        only $\loss{supcon}$ & & \cmark &  & \cmark &  & 6.2\std{1.1} & 2.85\std{0.66} \\ 
      \bottomrule
    \end{tabular}
\end{table}

\begin{table}[ht!]
    \centering
    \renewcommand{\arraystretch}{1.2}
    \begin{adjustbox}{max width=\columnwidth}
    \begin{tabular}{r|cc>{\columncolor{Gray}}ccc}
      \toprule
       $K$ & 5 & 8 & 10 & 12 & 16 \\
       & & $=\lvert\mathcal{C}\rvert$ & & & $=2\lvert\mathcal{C}\rvert$ \\
       \hline
       RoDeO & 25.0\std{1.0} & 21.7\std{0.8} & \textbf{34.0\std{1.3}} & 23.8\std{0.9} & 23.6\std{0.9} \\
       AP@0.3 & 6.88\std{0.80} & 5.40\std{0.78} & \textbf{9.44\std{0.90}} & 7.01\std{0.80} & 5.53\std{0.69} \\
      \bottomrule
    \end{tabular}
    \end{adjustbox}
    \caption{Ablation study on the number of ROI tokens $K$.}
    \label{tab:ablation_K}
\end{table}

\begin{table}[ht!]
    \centering
    \renewcommand{\arraystretch}{1.2}
    \begin{tabular}{r|>{\columncolor{Gray}}cccc}
      \toprule
       $\beta$ & 2.0 (normal) & 3.0 & 4.0 & 5.0 \\
       \hline
       RoDeO & \textbf{34.0\std{1.3}} & 32.2\std{1.3} & 31.9\std{1.3} & 30.5\std{1.1} \\
       AP@0.3 & \textbf{9.44\std{0.90}} & 8.55\std{0.87} & 8.07\std{0.91} & 8.61\std{0.91} \\
      \bottomrule
    \end{tabular}
    \caption{Ablation study on the shape parameter $\beta$ of the generalized Gaussian distribution.}
    \label{tab:ablation_beta}
\end{table}

\begin{table}[ht]
    \centering
    \caption{Ablation study on the usage of the no finding class $\nf$ during MIL aggregation (and in the BCE losses $\lossp{bce}$ and $\lossr{bce}$). We experimented with ignoring it and using only the classes in $\mathcal{C}$, or additionally using either the AND-aggregation ($\nfand$) or the OR-aggregation ($\nfor$) of the no-finding class but found that using both of them (marked in grey) is most effective for both BCE losses.}
    \label{tab:ablation_bce_nofind}
    \renewcommand{\arraystretch}{1.2}
    \begin{tabular}{lcccc}
        \toprule
        & \multicolumn{2}{c}{$\lossp{bce}$} & \multicolumn{2}{c}{$\lossr{bce}$} \\
        \cmidrule(lr){2-3}\cmidrule(lr){4-5}
        & RoDeO & AP@0.3 & RoDeO & AP@0.3 \\
        \midrule
        $\mathcal{C}$ & 18.0\std{0.7} & 3.95\std{0.51} & 32.3\std{1.3} & 9.03\std{0.99} \\ 
        $\mathcal{C}\cup\{\nfand\}$ & 23.2\std{0.9} & 3.05\std{0.69} & 33.3\std{1.2} & 8.75\std{0.92} \\ 
        $\mathcal{C}\cup\{\nfor\}$ & 22.8\std{0.8} & 5.05\std{0.56} & 32.3\std{1.3} & 8.42\std{0.89} \\ 
        $\mathcal{C}\cup\{\nfand, \nfor\}$ & \cellcolor{Gray}\textbf{34.0\std{1.3}} & \cellcolor{Gray}\textbf{9.44\std{0.90}} & \cellcolor{Gray}\textbf{34.0\std{1.3}} & \cellcolor{Gray}\textbf{9.44\std{0.90}} \\
        \bottomrule
    \end{tabular}
\end{table}

\begin{table}[ht]
    \centering
    \caption{Ablation study on different patch sizes. We experimented with two options to reduce the patch size from $32 \times 32$ to $16 \times 16$: (a) using lower-level features from \textit{denseblock3} instead of \textit{denseblock4}, and (b) skipping the last pooling features. For both cases, we observe no significant differences in performance.}
    \label{tab:ablation_numpatch}
    \renewcommand{\arraystretch}{1.2}
    \begin{tabular}{lcc}
        \toprule
        & RoDeO & AP@0.3 \\
        \midrule
        \rowcolor{Gray} default ($32 \times 32$) & \textbf{34.0\std{1.3}} & \textbf{9.44\std{0.90}}\\
        denseblock3 ($16 \times 16$) & 32.7\std{1.3} & 9.12\std{0.82}\\
        skip last pooling layer ($16 \times 16$) & 31.9\std{1.3} & 8.95\std{0.86} \\
        \bottomrule
    \end{tabular}
\end{table}

We conduct extensive ablation studies to quantify the relevance of different loss functions (\cref{tab:ablation_loss}), the influence of the number of ROI tokens $K$ (\cref{tab:ablation_K}), the treatment of the no-finding class (\cref{tab:ablation_bce_nofind}), the assumed distribution of the soft receptive field (\cref{tab:ablation_beta}), and the patch size (\cref{tab:ablation_numpatch}).

\paragraph{Loss functions}
In \cref{tab:ablation_loss}, we observe that without the patch branch loss components ($\lossp{} = \lossp{bce} + \lossp{supcon}$), the performance drops substantially in both RoDeO and AP, highlighting the importance of the patch branch for stabilizing the training. If, instead, the ROI branch loss components ($\lossr{} = \lossr{bce} + \lossr{supcon}$) are removed, the performance drops as well, but the model is still competitive with the best baselines. Here, the consistency loss $\losspr$ trains the ROI branch based on the predicted patch classes. Training without the consistency loss $\losspr$ leads to poor performance, again confirming the relevance of the consistency loss for stabilizing the box predictions based on the patch branch.

Additionally, we study the relevance of the BCE ($\lossp{bce}$, $\lossr{bce}$) and supervised contrastive ($\lossp{supcon}$, $\lossr{supcon}$) loss functions. Removing the BCE losses always leads to a performance collapse, independently of using the consistency loss $\losspr$. 
Removing the supervised contrastive losses leads to a performance drop, but the performance does not collapse entirely if the consistency loss is used.

\paragraph{Number of ROI tokens}
The number of ROI tokens $K$ determines the maximum number of proposed boxes and is a crucial parameter of our proposed method.
\cref{tab:ablation_K} shows the detection performance for varying values for this parameter.
The optimum is found at $K=10$ ($=\lvert\mathcal{C}\rvert + 2$) tokens.
A notably anomaly is at $K=8$ ($=\lvert\mathcal{C}\rvert$) tokens.
We hypothesize that in this case, the single tokens can get too class-specific, which hurts performance.

\paragraph{No-finding handling}
In \cref{tab:ablation_bce_nofind}, we study the use of the no-finding class $\nf$ during MIL-aggregation (and in the BCE losses $\lossp{bce}$ and $\lossr{bce}$). While in our standard setting, we use both \texttt{OR} ($\nfor$) and \texttt{AND} ($\nfand$) interpretations of this class, we also experiment with using only one of them and ignoring the no-finding class completely (considering only the classes in $\mathcal{C}$).
While this hyperparameter has minimal influence on the ROI branch, and the other settings are still competitive with the baselines, the performance degrades when changing this hyperparameter for the patch branch.

\paragraph{Receptive field distribution}
We assume that the relevant features for a pathology roughly follow a Gaussian distribution.
We check the validity of this assumption by gradually switching to a more \say{box-like} distribution by increasing the parameter $\beta$ of the generalized Gaussian distribution.
\cref{tab:ablation_beta} shows that performance is maximal at $\beta = 2$ which corresponds to a standard Gaussian distribution.

\paragraph{Patch size}
\cref{tab:ablation_numpatch} shows results for changing the patch size of the encoder from $32 \times 32$ to $16 \times 16$.
We employed two distinct strategies to investigate the influence of the patch size: First, we used the features of \textit{denseblock3} instead of \textit{denseblock4} in the DenseNet121 encoder. This feature map contains twice as many patches with half the size. Since the features of \textit{denseblock3} may differ significantly from those of \textit{denseblock3}, we experimented with another technique of skipping the final average pooling layer. This results in the same patch size, but the features are closer to those of our proposed models. Neither alternative improved over our default model.

\paragraph{Single- vs multi-box}
Lastly, we measure how our model performs on images with a single (single-box) and multiple target classes (multi-box).
In the former, WSRPN achieves a high RoDeO score of $36.4 \pm 1.6$, while for the latter, more difficult cases, the score drops to $23.7 \pm 1.5$.

%% file: sections/05_discussion.tex
\section{Discussion and Conclusion} \label{sec:discussion}
\subsection{Clinical applicability}
Our model WSRPN shows promising results for pathology localization on chest X-rays. It can provide precise or rough localization for most of the studied pathologies, even if bounding boxes are sometimes too huge. In clinical practice, even such rough localizations can provide massive value as they can assist clinicians in quickly spotting pathologies, especially in time-critical situations like emergency units. 
However, we also found some limitations that restrict its current clinical applicability. Most importantly, it often misclassifies some of the pathologies.
Note that the risks of misclassification differ between pathologies. For example, misclassifying a mass as a nodule does not have severe consequences as one of them is detected since both are indicators of cancer and require further examination.
Misclassifying (or missing) pneumothorax, on the other hand, is more critical as immediate clinical intervention may be required.
Therefore, future work may focus on improving the classification capabilities of our WSRPN model. 

\subsection{A novel approach towards weakly supervised pathology detection}
We propose the first WSup-OD method that can directly optimize (i.e., is differentiable w.r.t.) the box parameters (position and size). 
Existing WSup-OD methods rely on unsupervised, non-differentiable region proposals (MIL-based methods) or predict bounding boxes using thresholding (CAM-based methods). 
On the other hand, our Gaussian ROI pooling enables the box parameters to be optimized directly by different kinds of supervision signals, even simultaneously, which is impossible with current other approaches.
This enables a wide range of applications beyond WSup-OD, including, but not limited to, the integration into multimodal large language models, contrastive learning with text, or semi-supervised learning with bounding boxes for a subset of samples. 
We are convinced that -- besides setting a new state-of-the-art on this challenging task -- we open up a new research direction without the need for thresholding or external box proposals, which enables this underexplored field (of weakly supervised pathology detection) to progress beyond the existing approaches on which research has mostly stagnated in recent years.

\subsection{Conclusion}
We have proposed WSRPN -- a new paradigm for WSup-OD using learned box proposals -- after identifying weaknesses in the established box proposal algorithms when applied to X-ray images.
While further clinical validation is required, we set a new state-of-the-art in disease detection on the challenging CXR8 \cite{cxr8} dataset and significantly improve upon existing methods.
MIL-based methods for natural images have improved dramatically over several years, and we expect a similar evolution for RPN-MIL methods. We deem incorporating other forms of weak supervision like text, anatomy information, or semi-supervision into our framework as promising future research.

%% file: nomenclature.tex
\nomenclature[01]{$\mathcal{P}$}{Patch branch}
\nomenclature[02]{$\mathcal{R}$}{ROI branch}
\nomenclature[03]{$H$}{Patch height}
\nomenclature[04]{$W$}{Patch width}
\nomenclature[05]{$\hp$}{Embedding features of patch $(m,n)$}
\nomenclature[06]{$\hat{\bm{h}}^\mathcal{R}_{k}$}{Token features of token $k$}
\nomenclature[07]{$\hr$}{Embedding features for ROI $k$}
\nomenclature[08]{$A_{k,m,n}$}{Soft receptive field of token $k$ at patch $(m,n)$}
\nomenclature[09]{$p^{\mathcal{R} \rightarrow \mathcal{P}}_{m, n, c}$}{Spatial class map for patch $(m,n)$ and class $c$}
\nomenclature[10]{$\agghpc$}{Per-class patch features}
\nomenclature[11]{$\agghrc$}{Per-class ROI features}
\nomenclature[12]{$\tilde{p}^\mathcal{P}_{m, n, c}$}{Logits for patch $(m,n)$ and class $c$}
\nomenclature[13]{$\ppc$}{Probabilities for patch $(m,n)$ and class $c$}
\nomenclature[14]{$\prc$}{Probabilities for ROI $k$ and class $c$}
\nomenclature[15]{$\aggppc$}{Aggregated patch probabilities for class $c$}
\nomenclature[16]{$\aggprc$}{Aggregated ROI probabilities for class $c$}

\nomenclature[17]{$\bm{\mu}_k$}{Center coordinates of box $k$}
\nomenclature[18]{$\bm{\sigma}_k$}{Size of box $k$}

\nomenclature[19]{$\mathcal{C}$}{Set of classes}
\nomenclature[20]{$\nf$}{No-finding class}
\nomenclature[21]{$\phi$}{\emph{Sigmoid} function}
\nomenclature[22]{$\lse$}{\emph{LogSumExp} function}
\nomenclature[23]{$\mathcal{L}$}{Loss}
